\documentclass[journal]{IEEEtran}
\usepackage{algorithm}
\usepackage{algpseudocode}
\usepackage{algorithmicx}
\usepackage{multirow}
\usepackage{tabularx}
\usepackage{graphicx}
\usepackage{enumitem}
\usepackage{epstopdf}
\usepackage{ctable}
\usepackage{amsmath}
\usepackage{amsthm}
\usepackage{lineno}

\restylefloat{table}
\usepackage{array}

\usepackage{fancyhdr}
\usepackage{lastpage}
\begin{document}
\title{Novel Approaches for Predicting Risk Factors of Atherosclerosis}
\author{V. Sree Hari Rao, ~\IEEEmembership{Senior Member,~IEEE} and M. Naresh Kumar
\thanks{Manuscript received July 24, 2012; revised October 19, 2012; accepted
October 27, 2012. Date of publication; Date of current version. This work
was supported by the Foundation for Scientific Research and Technological Innovation
(FSRTI), a constituent division of Sri Vadrevu Seshagiri Rao Memorial
Charitable Trust, Hyderabad, India. The authors would like to thank the associate
editor and the anonymous reviewers for their constructive suggestions.}
\thanks{M. Naresh Kumar, National Remote Sensing Center (ISRO),  Hyderabad, Andhra Pradesh, $500 625$, India. E-mail: nareshkumar\_m@nrsc.gov.in}
\thanks{V. Sree Hari Rao, Institute for Development and Research in Banking Technology, Masab Tank, Hyderabad, Andhra Pradesh, $500 057$, India. E-mail: vshrao@idrbt.ac.in}
\thanks{Color versions of one or more of the figures in this paper are available online at http://ieeexplore.ieee.org.}
\thanks{\scriptsize $\copyright$ 20xx IEEE. Personal use of this material is permitted. Permission from IEEE must be obtained for all other uses, in any current or future media, including reprinting/republishing this material for advertising or promotional purposes, creating new collective works, for resale or redistribution to servers or lists, or reuse of any copyrighted component of this work in other works. DOI:  10.1109/TITB.2012.2227271}
}
\maketitle
\markboth{\scriptsize $\copyright$ 20xx IEEE. Personal use of this material is permitted. Published in IEEE.  DOI: 10.1109/TITB.2012.2227271}{\scriptsize \thepage}
\begin{abstract}
Coronary heart disease (CHD) caused by hardening of artery walls due to cholesterol known as atherosclerosis is responsible for large number of deaths world-wide. The disease progression is slow, asymptomatic and may lead to sudden cardiac arrest, stroke or myocardial infraction. Presently, imaging techniques are being employed to understand the molecular and metabolic activity of atherosclerotic plaques to estimate the risk. Though imaging methods are able to provide some information on plaque metabolism they lack the required resolution and sensitivity for detection. In this paper we consider the clinical observations and habits of individuals for predicting the risk factors of CHD. The identification of risk factors helps in stratifying patients for further intensive tests such as nuclear  imaging or coronary angiography. We present a novel approach for predicting the risk factors of atherosclerosis with an in-built imputation algorithm and particle swarm optimization (PSO). We compare the performance of our methodology with other machine learning techniques on STULONG dataset which is based on longitudinal study of middle aged individuals lasting for twenty years. Our methodology powered by PSO search has identified physical inactivity as one of the risk factor for the onset of atherosclerosis in addition to other already known factors. The decision rules extracted by our methodology are able to predict the risk factors with an accuracy of $99.73\%$ which is higher than the accuracies obtained by application of the state-of-the-art machine learning techniques presently being employed in the identification of atherosclerosis risk studies.
\end{abstract}
\begin{IEEEkeywords}
Atherosclerosis, Classification, Risk factors, Prediction, Imputation, Feature selection, Particle swarm optimization, Decision trees
\end{IEEEkeywords}
\section{Introduction}
\IEEEPARstart{A}{therosclerosis} is a systemic/chronic disease characterized by the accumulation of inflammatory cells and lipids in the inner lining of the arteries. It is a leading cardiovascular disease causing deaths worldwide~\cite{Farouc2006,Christopher1997}. Four million Americans have survived a stroke and lead disabled lives. More than $1$ out of $3$ ($83$ million) U.S. adults currently live with one or more types of cardiovascular diseases~\cite{CDC2011}. In $2010$, the total amount spent on cardiovascular diseases in the United States was estimated to be $\$444$ billion~\cite{CDC2011}. The number of deaths due to coronary artery disease in India were projected to increase from $1.591$ million in the year $2000$ to $2.034$ million by the year $2010$~\cite{Bhanot1999}.

Atherosclerosis has a long asymptomatic phase with a sub-clinical incubation period ranging from $30$ to $50$ years. The physicians would like to assess the risk of the patients having a severe clinical event such as stroke or heart attack and predict the same if possible. Some of the major known risk factors that eventually lead to the development of atherosclerosis are as follows: (i) family history of premature coronary heart disease or stoke in a first degree relative under the age of $60$, (ii) tobacco abuse, (iii) type II diabetes, (iv) high blood pressure, (v) left ventricular hypertrophy, (vi) high triglycerides, (vii) high low-density lipoprotein (LDL) cholesterol (viii) low high-density lipoprotein (HDL) cholesterol, (ix) high total cholesterol. Large number of factors influence the onset of atherosclerosis making it a difficult task for the physicians to diagnose in its early stages. Though main risk factors are identified, development of automated effective risk prediction models using data mining techniques becomes essential for better health monitoring and prevention of deaths due to cardiovascular diseases~\cite{Kukar1999,luckas2002}. Establishing a diagnostic procedure for early detection of atherosclerosis disease is very important as any delay would increase the risk of serious complications or even disability. Determining the conditions (risk factors) predisposing the development of atherosclerosis can lead to tests for identifying the disease in its early stages.

Though the clinical risk factors of CHD are identified there is a need for additional understanding on the disease progression for effective management~\cite{Natasha2011}. Researchers have been focusing on techniques for quantifying atherosclerosis plaque morphology, composition, mechanical forces etc., hoping for better patient screening procedures. Imaging of atherosclerotic plaques helps in both diagnosis and monitoring of the progression for future management~\cite{Davies2004,Tardif2011}. There are two modes of imaging atherosclerosis (i) invasive and (ii) non-invasive. Among the invasive methods X-ray angiography is the gold standard imaging technique even though it has certain limitations in providing information on plaque composition~\cite{Liang2011}. The invasive procedures such as intravascular ultrasound (IVUS)~\cite{Garcia-Garcia2010} and angioscopy~\cite{Ueda2003} help in understanding the plaque size and to a limited extent its composition. The intravascular thermography~\cite{Belardi2005} aids in monitoring the changes in plaque composition and metabolism. The non-invasive procedures such as B-mode ultrasound~\cite{Magnussen2011}, computerized tomography (CT)~\cite{Wintermark2008} and magnetic resonance imaging (MRI)~\cite{Corti19042011}, can provide information on plaque composition on vascular beds but they fail to throw much light on the metabolic activity of the plaque inflammatory cells. Though nuclear imaging techniques such as single photon emission computed tomography (SPECT) and positron emission tomography (PET)~\cite{Laufer2009} have the potential for 2D and 3D surface reconstruction of thrombus using radio labels to provide information on molecular, cellular and metabolic activity of plaques~\cite{Choudhury2009}, they lack the required resolution, sensitivity for detection and functional assessment in medium to small size arteries found in coronary circulation.

Keeping in view the limitations of the imaging techniques there is still a greater need for developing automated methods for predicting the risk factors of atherosclerosis disease in individuals which would be of great help in reducing the disease related deaths. Machine learning approaches have been employed in a variety of real world problems to extract knowledge from data for predictive tasks. The presence of large number of attributes in medical databases affects the decision making process as some of the factors may be redundant or irrelevant. Also, the presence of missing values and highly skewed value distributions in the attributes of medical datasets require development of new preprocessing strategies. Feature selection methods are aimed at identifying feature subsets to construct models that can best describe the dataset. The other advantages in using feature selection methods include: identifying and removing of redundant/irrelevant features~\cite{liu2005}, reducing the dimensionality of the dataset, and improving the predictive capability of the classifier. The present study attempts to identify risk factors causing atherosclerosis and the possible risk that the individuals are running at.

In view of the above challenges, we present the following novel features of our work:
\begin{itemize}
\item identifying the missing values (MV) in the dataset and imputing them by using a newly developed non-parametric imputation procedure;
\item determining factors that would help in predicting the risk of developing atherosclerosis among the different groups in the community;
\item building a predictive model that has the capability of rendering effective prediction of risk factors in realtime.
\item comparing the performance of our methodology with other state-of-the-art methods used in the identification of risk factors of atherosclerosis;
\item estimating the time complexity and scalability of our new methodology.
\end{itemize}
This paper is organized as follows: A brief survey of the state-of-the-art techniques employed in predicting the risk factors and understanding the biology of atherosclerosis is discussed in Section~\ref{review}, while in Section~\ref{chap8:sec1} we discuss a novel methodology for predicting the clinical risk factors of atherosclerosis. The description of the datasets, experiments and results are presented in Section~\ref{chap8:sec3}. The conclusions are placed in Section~\ref{chap8:sec6} and the discussion is deferred to Section~\ref{chap8:sec7}.
\section{A brief Survey of state-of-the-art techniques}
\label{review}
In~\cite{Luckas2003} a test for predicting atherosclerosis is proposed using genetic algorithm and a fitness function that depends on area under the curve (AUC) of receiver operator characteristics (ROC). In~\cite{Couturier2004} a three step approach based on clustering, supervised classification and frequent itemsets search is adopted to predict if a patient can develop atherosclerosis according to the correlation between his or her habits and the social environment. Support vector machines were employed in~\cite{Hongzong2007} for discriminating patients between coronary and non-coronary heart disease. Supervised classifiers such as Naive Bayes (NB), Multi Layer Perceptron (MLP), Decision Trees (DT) utilize the associations among the attributes for predicting future cardiovascular disorders in the individuals~\cite{Jose2006}. A correlation based feature selection with C4.5 decision tree is applied~\cite{Tsang-Hsiang2006} for risk prediction of cardiovascular disease. Recent developments in imaging methods for diagnosis have given new insights on the molecular and metabolic activity of atherosclerotic plaques~\cite{Crouse2006,Corti19042011,Ibanez2009}. There are many other studies wherein machine learning techniques have also been employed for predicting the risk of CHD due to atherosclerosis using ultrasound and other imaging methods~\cite{Mougiakakou2007,Christodoulou2003}. Community based studies help in understanding the risk factors of atherosclerosis in different social strata~\cite{Chambless2003,Selvin2005}.

In the above studies the missing values (MV) were either deleted~\cite{Tsang-Hsiang2006}, or filled with approximate values~\cite{Luckas2003}. A windowing method was employed~\cite{Jiri2008} for obtaining the aggregates of the attributes for imputation of MVs. These approaches would lead to biased estimates and may either reduce or exaggerate the statistical power. Methods such as logistic regression, maximum likelihood and expectation maximization have been employed for imputation of MV, but they can be applied only on data sets that are either nominal or numeric. There are other imputation methods such as k-nearest neighbor imputation (KNNI)~\cite{BM2003}; k-means clustering imputation (KMI)~\cite{DSSL2004}; weighted k-nearest neighbor imputation (WKNNI)~\cite{TCSBHTBA2001} and fuzzy k-means clustering imputation (FKMI)~\cite{DSSL2004}.
\section{Novel Methodology for Predicting Risk factors of Atherosclerosis}
\label{chap8:sec1}
The mean value imputation proposed by Sree Hari Rao and Naresh Kumar~\cite{Rao2011} can be employed only when the attribute values are normally distributed. In case of highly skewed attribute values the above method may result in biased estimates as mean value is not a true representative of a non-normal distribution. Motivated by the above issues we propose a methodology comprising of a novel nonparametric missing value imputation method that can be applied on (i) data sets consisting of attributes that are of the type categorical (nominal) and/or numeric (integer or real), and (ii) attribute values that belong to highly skewed distribution. The methodology proposed by Freud and Mason~\cite{Freund1999} ignores missing values while generating the decision tree, which renders lower prediction accuracies.
In this paper we propose a new feature subset selection methodology where in, a particle optimization search (PSO) is wrapped around an alternating decision tree (ADT) embedded with new imputation strategy discussed in~Section~\ref{chap9:RNI}, for generation of effective decision rules. This methodology can predict the diagnosis of CHD in real time. In fact the decision rules obtained by employing this novel methodology will be useful to diagnose other individuals based on their risk factors. We designate the present machine learning approach as predictive risk assessment of atherosclerosis ($PRAA$) methodology throughout this work.

We adopt the procedure suggested by Sree Hari Rao and Naresh Kumar~\cite{Rao2011} for building and evaluating the classification model using an ADT.
\subsection{Data Representation}
A medical dataset can be represented as a set \emph{S} having row vectors  $(R_{1},R_{2},\hdots,R_{m})$ and column vectors $(C_{1},C_{2},\hdots,C_{n})$. Each record can be represented as an ordered n-tuple of clinical and laboratory attributes $(A_{i1}, A_{i2},\hdots, A_{i(n-1)}, A_{in})$ for each $\emph{i}=1,2,\hdots,m$ where the last attribute $(A_{in})$ for each \emph{i}, represents the physician's diagnosis to which the record $(A_{i1}, A_{i2},\hdots, A_{i(n-1)})$ belongs and without loss of generality we assume that there are no missing elements in this set. Each attribute of an element in \emph{S} that is $A_{ij}$ for $\emph{i}=1,2,\hdots,m$ and $\emph{j}=1,2,\hdots,n-1$ can either be a categorical (nominal) or numeric (real or integer) type. Clearly all the sets considered are finite sets.
\subsection{New Imputation Strategy}
\label{chap9:RNI}
The first step in any imputation algorithm is to compute the proximity measure in the feature space among the clinical records to identify the nearest neighbors from where the values can be imputed. The most popular metric for quantifying the similarity between any two records is the Euclidean distance. Though this metric is simpler to compute, it is sensitive to the scales of the features involved. Further it does not account for correlation among the features. Also, the categorical variables can only be quantified by counting measures which calls for the development of effective strategies for computing the similarity~\cite{Tadashi2009}. Considering these factors we first propose a new indexing measure $I_{C_{l}}(R_{i},R_{k})$ between two typical elements $R_{i}$, $R_{k}$ for $i,k=1,2,\ldots,m,$ $l=1,2,\ldots,n-1$ belonging to the column $C_{l}$ of \emph{S} which can be applied on any type of data, be it categorical (nominal) and/or numeric (real). We consider the following cases:
\begin{description}
  \item[Case I:]~ $A_{in}=A_{kn}$ \\
Let \emph{A} denote the collection of all members of \emph{S} that belong to the same decision class to which $R_{i}$ and $R_{k}$ belong. Based on the type of the attribute to which the column $C_{l}$ belongs, the following situations arise:
\begin{itemize}
  \item [(i)]
  Members of the column $C_{l}$ of S i.e
  $(A_{1l},A_{2l},\ldots,A_{ml})^{T}$ are of nominal or categorical or integer type:\\
\noindent We now express \emph{A} as a disjoint union of non-empty subsets of \emph{A}, say $B_{\gamma_{p_{1l}}},B_{\gamma_{p_{2l}}},\ldots,B_{\gamma_{p_{sl}}}$ obtained in such a manner that every element of \emph{A} belongs to one of these subsets and no element of \emph{A} is a member of more than one subset of \emph{A}. That is $\emph{A}=B_{\gamma_{p_{1l}}}\bigcup B_{\gamma_{p_{2l}}}\bigcup, \ldots,\bigcup B_{\gamma_{p_{sl}}}$, in which $\gamma_{p_{1l}}, \gamma_{p_{2l}}, \ldots, \gamma_{p_{sl}}$  denote the cardinalities of the  respective subsets $B_{\gamma_{p_{1l}}},B_{\gamma_{p_{2l}}},\ldots,B_{\gamma_{p_{sl}}}$ formed out of the set \emph{A}, with the property that each member of the same subset has the same first co-ordinate and members of no two different subsets have the same first co-ordinate.  We define an index $I_{C_{l}}(R_{i},R_{k})$ for each $l=1,2,\ldots,n-1$
\begin{eqnarray*}
    I_{C_{l}}(R_{i},R_{k})=\left\{
                             \begin{array}{ll}
                              \min\{\frac{\gamma_{p_{il}}}{\gamma},\frac{\gamma_{q_{kl}}}{\gamma}\}, & \hbox{for $i \neq k$;} \\
                               0, & \hbox{otherwise.}
                             \end{array}
                           \right.
\end{eqnarray*}
\noindent where $\gamma_{p_{il}}$ represents the cardinality of the subset $B_{\gamma_{p_{il}}}$, all of whose elements have first co-ordinates $A_{il}$, $\gamma_{q_{kl}}$ represents the cardinality of that subset $B_{\gamma_{q_{kl}}}$, all of whose elements have first co-ordinates $A_{kl}$ and $\gamma=\gamma_{p_{1l}}+ \gamma_{p_{2l}}+ \ldots,+\gamma_{p_{sl}}$ represents the cardinality of the set $A$.
\item [(ii)] Members of the column $C_{l}$ of S i.e $(A_{1l},A_{2l},\ldots,A_{ml})^{T}$   are of real (fractional or non-integer numbers):\\
\noindent We consider the set $P_{l}$ for $l=1,2,\dots,n-1$ which is a collection of all the members of the column $C_{l}$. We then compute the skewness measure $sk(P_{l})=\frac{\frac{1}{m}\sum_{i=1}^{m}(A_{il}-\overline{A_{il}})^3}{(\sqrt \frac{1}{m} \sum_{i=1}^{m}(A_{il}-\overline{A_{il}})^2)^3}$ where $ \overline{A_{il}}$ denote the mean of $A_{il}$  for each $l=1,2,\ldots,n-1$.
Define the sets $\begin{array}{@{\hspace{0mm}}r@{\;}l@{\hspace{0mm}}}        M_{l} & = \displaystyle \left\{a\in P_{l}|a \leq A_{il}, \hbox{for $sk(P_{l}) < 0$} \right\}\end{array}$ or, $\begin{array}{@{\hspace{0mm}}r@{\;}l@{\hspace{0mm}}}
M_{l} & = \displaystyle \left\{b\in P_{l}|b > A_{il}, \hbox{for $sk(P_{l}) \geq 0$}\right\}\end{array}$ and similarly $\begin{array}{@{\hspace{0mm}}r@{\;}l@{\hspace{0mm}}}
N_{l} & = \displaystyle \left\{a\in P_{l}|a \leq A_{kl},\hbox{for $sk(P_{l}) < 0$}  \right\}\end{array}$ or, $\begin{array}{@{\hspace{0mm}}r@{\;}l@{\hspace{0mm}}}
N_{l} & = \displaystyle \left\{b\in P_{l}|b > A_{kl},\hbox{for $sk(P_{l}) \geq 0$}  \right\}\end{array}$. Let $\tau_{l}$ and $\rho_{l}$ be the cardinalities of the sets $M_{l}$ and $N_{l}$ respectively. Construct the index $I_{C_{l}}(R_{i},R_{k})$,
\begin{eqnarray*}
I_{C_{l}}(R_{i},R_{k})=\left\{
                             \begin{array}{ll}
                             \min\{\frac{\tau_{l}}{\gamma},\frac{\rho_{l}}{\gamma}\}, & \hbox{for $i \neq k$;} \\
                               0, & \hbox{otherwise.}
                             \end{array}
                           \right.
\end{eqnarray*}
\noindent In the above definition $\gamma$ represents the cardinality of the set $P_{l}$.
\end{itemize}
\item [Case II:]~ $A_{in}\neq A_{kn}$ \\
\label{case2}
Clearly $R_{i}$ and $R_{k}$ belong to two different decision classes. Consider the subsets $P_{i}$ and $Q_{k}$ consisting of members of $\emph{S}$ that share the same decision with $R_{i}$ and $R_{k}$ respectively. Clearly $P_{i}\bigcap Q_{k} = \emptyset$. Based on the type of the attribute of the members, the following situations arise:
\begin{itemize}
  \item[(i)] Members of the column $C_{l}$ of S i.e $(A_{1l},A_{2l},\ldots,A_{ml})^{T}$ are of nominal or categorical type:\\
 \noindent Following the procedure discussed in Case I item (i) we write \emph{$P_{l}$} and \emph{$Q_{l}$} for each $l=1,2,\ldots,n-1$ as a disjoint union of non-empty subsets of  $P_{\beta_{1l}},P_{\beta_{2l}},\ldots,P_{\beta_{rl}}$ and $Q_{\delta_{1l}},Q_{\delta_{2l}},\ldots,Q_{\delta_{sl}}$ respectively in which $\beta_{1l}, \beta_{2l}, \ldots, \beta_{rl}$  and $\delta_{1l}, \delta_{2l}, \ldots, \delta_{sl}$ indicate the cardinalities of the respective subsets. We define the indexing measure between the two records $R_{i}$ and $R_{k}$  as
\begin{eqnarray*}
    I_{C_{l}}(R_{i},R_{k})=\left\{
                             \begin{array}{ll}
                               \max\{\frac{\beta_{rl}}{\delta_{sl}+\beta_{rl}},\frac{\delta_{sl}}{\delta_{sl}+\beta_{rl}}\}, & \hbox{for $i \neq k$;} \\
                               0, & \hbox{otherwise.}
                             \end{array}
                           \right.
\end{eqnarray*}
where $\beta_{rl}$ represents the cardinality of the subset $P_{\beta_{rl}}$ all of whose elements have first co-ordinates $A_{il}$ in set $P_{l}$ and $\delta_{sl}$ represents the cardinality of that subset $Q_{\delta_{sl}}$, all of whose elements have first co-ordinates $A_{kl}$ in set $Q_{l}$. \label{case2:item1}
\item[(ii)] Members of the column $C_{l}$ of S i.e $(A_{1l},A_{2l},\ldots,A_{ml})^{T}$   are of numeric type:\\
\noindent  If the type of the attribute is an integer we follow the procedure discussed in Case II item (i) . For fractional numbers we follow the procedure discussed in Case I item (ii) and we define the set $P_{l}$ as the members of column $C_{l}$ that belong to decision class $A_{in}$ and $Q_{l}$ as the members of column $C_{l}$ that belong to decision class $A_{kn}$. We then compute the skewness measure $sk(P_{l})=\frac{\frac{1}{m}\sum_{i=1}^{m}(A_{il}-\overline{A_{il}})^3}{(\sqrt \frac{1}{m} \sum_{i=1}^{m}(A_{il}-\overline{A_{il}})^2)^3}$ where $ \overline{A_{il}}$ denote the mean of $A_{il}$  for each $l=1,2,\ldots,n-1$ and $sk(Q_{l})=\frac{\frac{1}{m}\sum_{i=1}^{m}(A_{kl}-\overline{A_{kl}})^3}{(\sqrt \frac{1}{m} \sum_{i=1}^{m}(A_{kl}-\overline{A_{kl}})^2)^3}$ where $ \overline{A_{kl}}$ denote the mean of $A_{kl}$  for each $l=1,2,\ldots,n-1$.
We then construct the sets $T_{l}$ and $S_{l}$ as follows:
$\begin{array}{@{\hspace{0mm}}r@{\;}l@{\hspace{0mm}}}        T_{l} & = \displaystyle \left\{a\in P_{l}|a \leq A_{il}, \hbox{for $sk(P_{l}) < 0$} \right\}\end{array}$ or, $\begin{array}{@{\hspace{0mm}}r@{\;}l@{\hspace{0mm}}}
T_{l} & = \displaystyle \left\{b\in P_{l}|b > A_{il}, \hbox{for $sk(P_{l}) \geq 0$}\right\}\end{array}$ and similarly $\begin{array}{@{\hspace{0mm}}r@{\;}l@{\hspace{0mm}}}
S_{l} & = \displaystyle \left\{a\in P_{l}|a \leq A_{kl},\hbox{for $sk(P_{l}) < 0$}  \right\}\end{array}$ or, $\begin{array}{@{\hspace{0mm}}r@{\;}l@{\hspace{0mm}}}
S_{l} & = \displaystyle \left\{b\in P_{l}|b > A_{kl},\hbox{for $sk(P_{l}) \geq 0$}  \right\}\end{array}$. We now define  the index $I_{C_{l}}(R_{i},R_{k})$ between the two records $R_{i},R_{k}$ as
\begin{eqnarray*}
    I_{C_{l}}(R_{i},R_{k})=\left\{
                               \begin{array}{ll}
                             \min\{\frac{\beta_{l}}{\lambda},\frac{\delta_{l}}{\lambda}\}, & \hbox{for $i \neq k$;} \\
                               0, & \hbox{otherwise.}
                             \end{array}
                           \right.
\end{eqnarray*}
\noindent In the above definition $\beta_{l}$ and $\delta_{l}$ represents the cardinalities of the sets $T_{l}$ and $S_{l}$ respectively. The sum of the cardinalities of the sets $P_{l}$ and $Q_{l}$ is represented by $\lambda$
\end{itemize}
\end{description}

The proximity or distance scores between the clinical records in the data set $S$ can be represented as $D=\{\{0,d_{12},\ldots,d_{1m}\};\{d_{21},0,\ldots,d_{2m}\};\ldots;\{d_{m1},d_{m2},\ldots,0\}\}$ where $d_{ik}=\sqrt{\sum_{l=1}^{n-1}I_{C_{l}}^{2}(R_{i},R_{k})}$. For each of the missing value instances in a record $R_{i}$ our imputation procedure first computes the score $\alpha(x_{k})=\frac{(x_{k}-median(x))}{median{|x_{i}-median(x)|}}$ where $\{x_{1},x_{2},\ldots,x_{n} \}$ denote the distances of $R$ from $R_{k}$. We then pick up only those records (nearest neighbors) which satisfy the condition $\alpha(x_{k}) \leq 0$ where $\{d_{i1},d_{i2},\ldots,d_{im} \}$ denote the distances of the current record $R_{i}$ to all other records in the data set $S$. If the type of attribute is categorical or integer, then the data value that has the highest frequency (mode) of occurrence in the corresponding columns of the nearest records is imputed. For the data values of type real we first collect all non zero elements in the set $D$ and denote this set by $B$. For each element in set $B$ we compute the quantity $\beta(j)$=$\frac{1}{B(j)}$ $\forall j=1,\ldots,\gamma$ where $\gamma$ denote the cardinality of the set $B$. We compute the weight matrix as $W(j)=\frac{\beta_{j}}{\sum_{i=1}^{\gamma} \beta(i)}$ $\forall j=1,\ldots,\gamma$. The value to be imputed may be taken as $\sum_{i=1}^{\gamma} P(i)*W(i)$.
\subsection{Particle swarm optimization search for feature subset selection (Risk factors)}
\label{FS}
 A PSO search consists of a set of particles initialized with a candidate solution to a problem. Each particle is associated with a position vector and a velocity vector. The particles evaluate the fitness of the solutions iteratively and store the location where they had their best fit known as the local best ($L$). The particles change their position and velocity iteratively in a suitable manner with respect to the best fit solution to reach a global optimal solution. The best fit solution among the particles is called the global best ($G$). We represent the position vector of the particle as a binary string and accuracy of the learning algorithm as the fitness function for evaluation. The velocity and position vectors of the particles are modified using the procedure suggested in~\cite{Kennedy2001}.

%
\begin{algorithm}[!t]\scriptsize
\caption{The $PRAA$ Methodology}
\label{chap9:PRA}
\begin{algorithmic}
\Require
\begin{enumerate}[label=(\alph{*})]
            \item Data sets for the purpose of decision making $S(m,n)$ where $m$ and $n$ are number of records and attributes respectively and the members of $S$ may have MV in any of the attributes except in the decision attribute, which is the last attribute in the record.
            \item  The type of attribute $C$ of the columns in the data set.
\end{enumerate}
\Ensure
\begin{enumerate}[label=(\alph{*})]
            \item  Classification accuracy for a given data set $S$.
            \item  Performance metrics AUC, SE, SP.
\end{enumerate}
   \textbf{Algorithm}
   \begin{enumerate}[label=(\arabic{*})]
        \item Identify and collect all records in a data  set $S$
        \item  Impute the MV in the data set $S$ using the procedure discussed in Section~\ref{chap9:RNI}.
        \item Extract the influential features using a wrapper based approach with  particle swarm optimization search for identifying feature subsets and ADT for its evaluation as discussed in Section~\ref{FS}.
       \item Split the dataset in to training and testing sets using a stratified $k$ fold cross validation procedure. Denote each training and testing data set by $T_{k}$ and $R_{k}$ respectively.
   \item For each $k$ compute the following
      \begin{enumerate}[label=(\roman{*})]\label{NRSproc}
        \item Build the ADT using the records obtained from $T_{k}$. \label{buildadt}
        \item Compute the predicted probabilities (scores) for both positive and negative diagnosis of CHD from the ADT built in Step (5)-(i) using the test data set $R_{k}$. Designate the set consisting of all these scores by $P$.
        \item Identify and collect the actual diagnosis from the test data set $R_{k}$ in to set denoted by $L$.\label{step3}
     \end{enumerate}
        \item  Repeat the Steps (5)-(i) to Step (5)-(iii) for each fold.
        \item Obtain the performance metrics AUC, SE and SP utilizing the sets $L$ and $P$.
         \item RETURN AUC, SE, SP.
         \item END.
   \end{enumerate}
\end{algorithmic}
\end{algorithm}
\section{Experiments and Results}
\label{chap8:sec3}
\subsection{Dataset}
In the present study we use the STULONG dataset~\cite{pkdd2004} which is a longitudinal primary preventive study of middle-aged men lasting twenty years for accessing the risk of atherosclerosis and cardiovascular health depending on personal and family history collected at Institute of Clinical and Experimental Medicine (IKEM) in Praha and the Medicine Faculty at Charles University in Plzen (Pilsen). The STULONG dataset is divided into four sub-groups namely Entry, Letter, Control and Death. The Entry dataset consists of $1417$ patient records with $64$ attributes having either codes or results of size measurements of different variables or results of transformations of the rest of the attributes during the first level examination. We utilize the Entry, Control and Death datasets for our predictive modeling. The Entry level dataset is divided into three groups (a) normal group (NG), (b) pathological group (PG), (c) risk group (RG), and (d) not allotted (NA) group, based on the studied group of patients (KONSKUP) in ($1$, $2$), ($5$), ($3$, $4$), ($6$) respectively. We form a new dataset by joining Entry, Control and Death datasets as follows: (i) we write the identification number of a patient (ICO) based on the selection criteria suggested in~\cite{Hoa2003} and determine the susceptibility of a patient to atherosclerosis based on the attributes recorded in Control and Death tables. An individual is considered to have cardiovascular disease if he or she has history of heart disease (i.e., he or she has at least one positive value on attributes such as myocardial infarction (HODN2), cerebrovascular accident (HODN3), myocardial ischaemia (HODN13), silent myocardial infarction (HODN14)), or died of heart disease (i.e., the record appears in the Death table with PRICUMR attribute equal to $05$ (myocardial infarction), $06$ (coronary heart disease), $07$ (stroke), or $17$ (general atherosclerosis)). Based on the above definition we divide the Entry dataset in to two datasets DS1 and DS2 depending on whether the patients are in NG or RG group respectively.

\subsection{Description of Experiments}
In our methodology we have employed a stratified ten-fold cross validation ( $k=10$) procedure. We have applied a standard implementation of SVM with radial basis function kernel using LibSVM package~\cite{chang2001}. We have taken the following standard parameter values for PSO (i) number of particles $Z=50$, (ii) number of iterations $G=100$, (iii) cognitive factor $c_1=2$, and (iv) social factor $c_2=2$.  The standard implementation of C4.5, Naive Bayes (NB), Multi Layer Perceptron (MLP) algorithms in Weka$^\copyright$~\cite{Witten2005} are considered for evaluating the performance of our algorithm. An implementation of correlation based feature selection (CFS)~\cite{Liu1996} algorithm with genetic search has been considered for comparing with our methodology. Also, we have implemented the $PRAA$ algorithm and the performance evaluation methods in Matlab$^\copyright$. A non-parametric statistical test proposed by Wilcoxon~\cite{Wilcoxon1945} is used to compare the performance of the algorithms.
\subsection{Performance Measures and Results}
The PRAA methodology has outperformed (see Table~\ref{chap8:tab1}) the classifiers C4.5, SVM, MLP and NB in terms of sensitivity (SE), specificity (SP) and AUC performance metrics. In risk group dataset DS2 our methodology could identify the patients with an accuracy of $99.73\%$ who are affected by atherosclerosis using only $13$ out of $51$ attributes. The wrapper based feature selection using PSO and ADT could identify  the influential factors such as alcohol (ALKOHOL), daily consumption of tea (CAJ), hypertension or ictus (ICT), hyperlipoproteinemia (HYPLIP), since how long hyper tension (HT) has appeared (HTTRV), before how many years  hyperlipidemia had appeared (HYPLTRV), blood pressure II systolic (DIAST2), cholesterol in mg \% (CHLST), Glucose in urine (MOC), obesity (OBEZRISK), hypertension (HTRISK) which are in conformity with other studies related to cardiovascular diseases~\cite{McGill2002,Drechsler2010,Lydia2003}. We have identified an important fact that even in normal group DS1 individuals who mostly confine to sitting positions without any physical activity (AKTPOZAM=1) may lead to atherosclerosis as observed in~\cite{Thompson2003}.

\begin{table}[!t]\scriptsize
\centering
\caption{Performance comparison of the $PRAA$ with other methodologies (C4.5, SVM, NB, MLP) on the data sets used in the present study}
\label{chap8:tab1}
\begin{tabular}{c||c||c||c||c||c}
\hline
\bf{Dataset}&\bf{Method}&\bf{Accuracy}&\bf{SE}&\bf{SP}&\bf{AUC} \\
&&(\%)&&&\\
\hline \hline
\multirow{6}{*}{DS1}&\bf $PRAA$&\bf 98.04& \bf 93.75& \bf 100.00& \bf 0.94\\
&C4.5& 70.59&  6.25&100.00&  1.00\\
&NB& 52.94& 56.25& 51.43&  0.53\\
&SVM& 35.29&100.00&  5.71&  0.53\\
&MLP& 66.67&  0.00& 97.14&  0.97\\
\hline
\multirow{6}{*}{DS2}&\bf $PRAA$&\bf 99.73& \bf 99.35&\bf 100.00&\bf 1.00\\
&C4.5& 50.00& 55.48& 45.97&  0.52\\
&NB& 61.20& 48.39& 70.62&  0.58\\
&SVM& 45.63& 92.26& 11.37&  0.56\\
&MLP& 57.92&  0.65&100.00&  1.00\\
\hline
\end{tabular}
\end{table}



\begin{figure}[!t]
\centering
\includegraphics[width=0.5\textwidth]{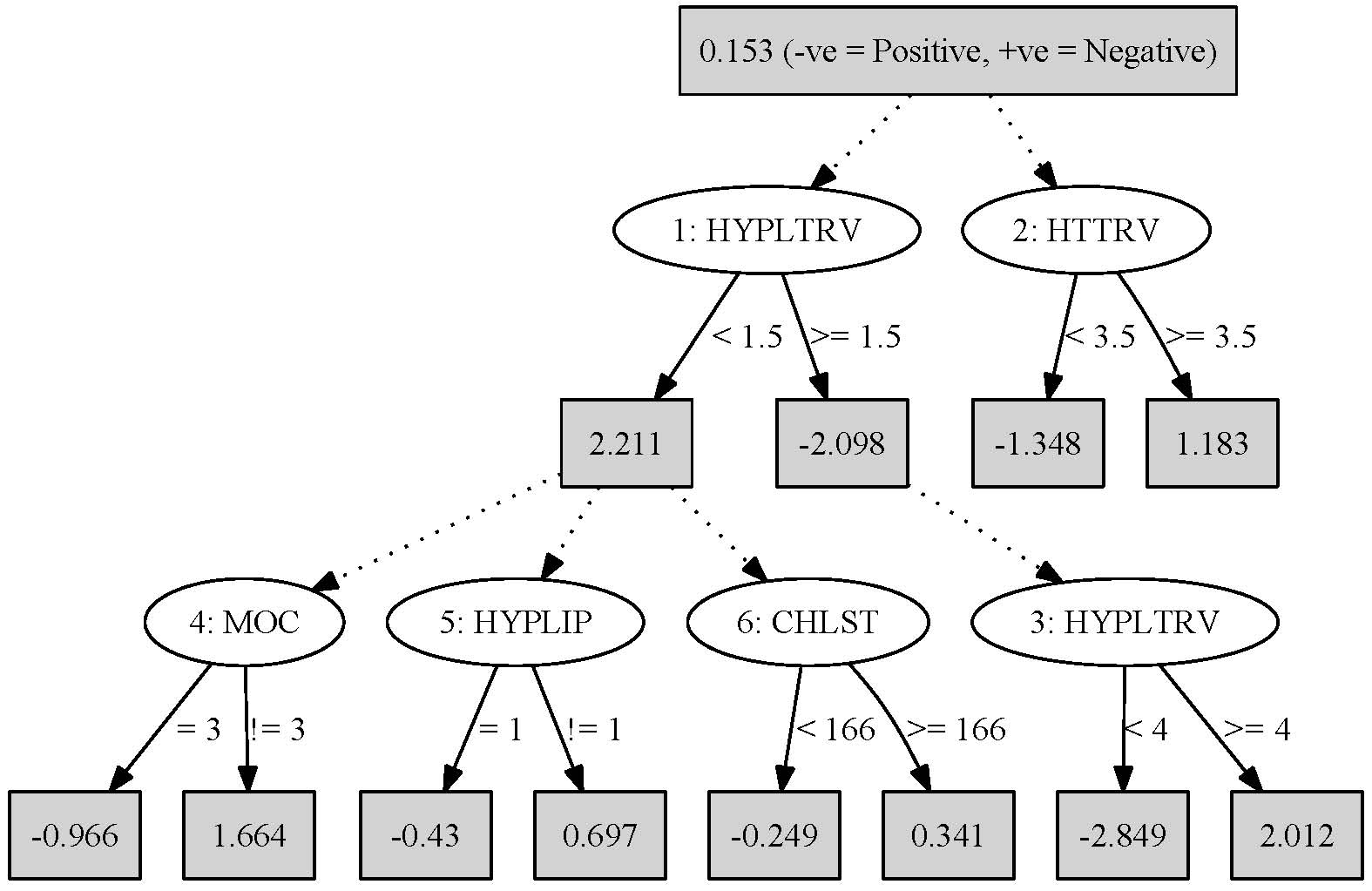}\\
  \caption{Alternating Decision tree generated for dataset DS2}\label{chap9:adt}
  \end{figure}

The ADT generated for the risk group (DS2) is shown in Fig.~\ref{chap9:adt}.
The following decision rules are extracted from the decision tree shown in Fig.~\ref{chap9:adt}:
\begin{enumerate}
\item The risk of atherosclerosis increases by a factor of  $-2.098$ if an individual is suffering from hyperlipidemia since $1.5$ years
 \item The presence of hypertension ($< 3.5$ years) would increase the risk of atherosclerosis by a factor of $-1.348$;
\item The risk is estimated as $-2.098 -1.348 - 2.849 -0.966-0.43-0.249=-7.94$ if an individual is suffering from hyperlipidemia from anywhere between $1.5$ years to $4$ years ($HYPLTRV \ge 1.5$ and $<4.0$), with hypertension since $3.5$ years and cholesterol levels less than $166$. It is observed that the presence of  hyperlipoproteinemia and Glucose in urine would increase the risk of atherosclerosis.
\end{enumerate}

%
%
\section{Conclusions}
\label{chap8:sec6}
A new methodology ($PRAA$) with built in features for imputation of missing values that can be applied on datasets wherein the attribute values are either normal and/or highly skewed having either categorical and/or numeric attributes and identification of risk factors using wrapper based feature selection is discussed. The $PRAA$ methodology has outperformed over the state-of-the-art methodologies in determining the risk factors associated with the onset of atherosclerosis disease. The $PRAA$ methodology has generated a decision tree with an accuracy of $99.7\%$ for dataset DS2. Based on the performance measures we conclude that the use of PSO search for feature subset selection wrapped around ADT embedded with new imputation strategy for fitness evaluation have improved the prediction accuracies.

\section{Discussion}
\label{chap8:sec7}
In this section we present a discussion on the performance of $PRAA$ methodology on benchmark datasets, its computational complexity, scalability and comparative studies with other methodologies.
\subsection{Performance Comparison of new imputation procedure on Benchmark Datasets}
\label{perMV}
Since no specific studies on imputation of missing values in cardiovascular disease data sets are available in the literature we have utilized some bench mark data sets obtained from Keel and University of California Irvin (UCI) machine learning data repositories~\cite{Alcala-Fdez2010,Frank2010} to test the performance of the new imputation algorithm. The Wilcoxon statistics in Table~\ref{chap8:tab5} is computed based on the accuracies obtained by the new imputation algorithm with the accuracies of those obtained by using embedded algorithms for handling missing values such as C4.5 decision tree. The results in Table~\ref{chap8:tab5} below clearly demonstrate that the imputation procedure presented in Section~\ref{chap9:RNI} has superior performance when compared to other imputation algorithms as the test statistics are well below or equal to the critical values with $p<0.05$ in all cases.
\begin{table}[H]\scriptsize
\caption{Wilcoxon sign rank statistics for matched pairs comparing the new imputation algorithm with other imputation methods using C4.5 decision tree}
\label {chap8:tab5}
\centering
\begin{tabular}{l||c||c||c||c}
\hline
\bfseries Method	&	\bfseries Rank Sums &\bfseries Test& \bfseries Critical& \bfseries p-value \\
& \bfseries (+, -)&\bfseries  Statistics&\bfseries Value&\\
\hline \hline
FKMI	&	28.0, 0.0&	0.0&	3	&0.02\\ \hline
KMI		&28.0, 0.0	&0.0	&3	&0.02\\ \hline
KNNI	&	15.0, 0.0&	0.0&0	&0.06\\ \hline
WKNNI	&	21.0, 0.0	&1.0	&18	&0.03\\ \hline
\end{tabular}
\end{table}
\subsection{Comparison with other Related Methodologies on Atherosclerosis}
\label{chap8:sec5}
In this section we compare the results obtained in~\cite{Kukar1999, Tsang-Hsiang2006,Hongzong2007} with the results of our new methodology on the risk group dataset DS2. As compared to~\cite{Tsang-Hsiang2006} where in CFS with genetic search and C4.5 is employed, an SE of $39.4\%$ and SP of $82.8\%$ is observed. In~\cite{Hongzong2007} SVM was used to classify the patients with an SE of $95\%$ and SP of $90\%$. In~\cite{Kukar1999}  both NB and MLP were used for classification with an accuracy of $80\%$ and the could obtain an SE of $92\%$ and $82\%$, an SP of $53\%$ and $76\%$ respectively. The new methodology when applied on the dataset DS2 resulted in an accuracy of $99.73\%$, an SE of $99.35\%$ and an SP of $100\%$ which is regarded as a good classification model since both SE and SP are higher than $80\%$.
\subsection{Computational Complexity and Scalability of $PRAA$ Algorithm}
\label{chap8:sec4}
\begin{figure}[!t]
\centering
\includegraphics[width=0.5\textwidth]{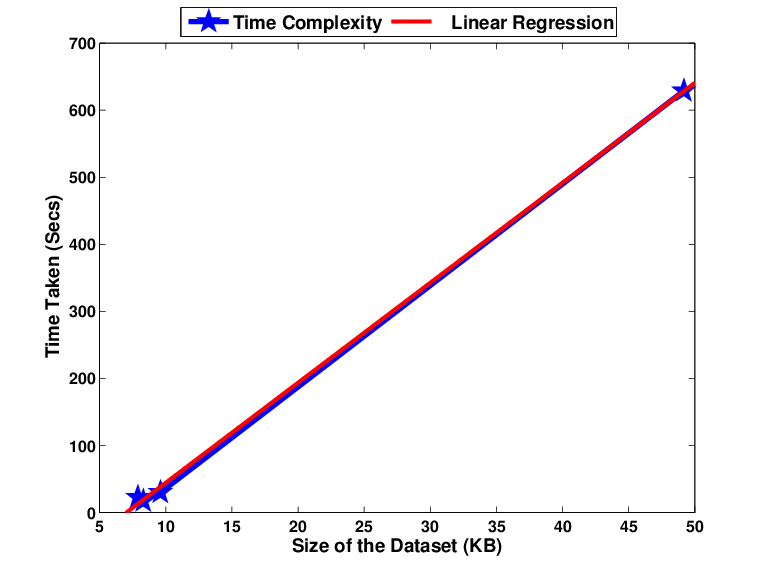}\\
  \caption{Computational complexity of the $PRAA$}\label{chap8:fig6}
\end{figure}

The computational complexity is a measure of the performance of the algorithm which can be measured in terms of the number of CPU clock cycles elapsed in seconds for performing the methodology on a dataset. For each data set having $n$ attributes and $m$ records, we select only those subset of records $m_{1} \le m$, in which missing values are present. The distances are computed for all attributes $n$ excluding the decision attribute. So, the time complexity for computing the distance would be $O(m_{1}*(n-1))$. The time complexity for computing skewness is $O(m_{1})$. The time complexity for selecting the nearest records is of order $O(m_{1})$. For computing the frequency of occurrences for nominal attributes and weighted average for numeric attributes the time taken would be of the order $O(m_{1})$. Therefore, for a given data set with $k$-fold cross validation having $n$ attributes and $m$ records, the time complexity of our new imputation algorithm would be $k*(O(m_{1}*(n-1)*m)+3*O(m_{1}))$ which is asymptotically linear. Our experiments were conducted on a personal computer having an Intel(R) core (TM) 2 Duo,  CPU @$2.93$ GHZ  processor with $4$ GB RAM and the time taken by $PRAA$ for varying database sizes is shown Fig.~\ref{chap8:fig6}. We have employed a linear regression on our results and obtained the relation between the time taken (T) and the data size (D) as $T=14.909D-104.655$, $\alpha=0.05$, $p=0.0003$, $r^2=0.994$.

The presence of the linear trend between the time taken and the varying database sizes ensure the numerical scalability of the performance of $PRAA$ methodology. A comparison with other related methodologies used in the study of atherosclerosis yields a conclusion that the new $PRAA$ methodology presented in this paper has a superior performance over other methods studied in~\cite{Kukar1999, Tsang-Hsiang2006,Hongzong2007}. We hold the view that more intensive and introspective studies of this kind will pave way for effective risk prediction and diagnosis of atherosclerosis.

\end{document}